\pdfoutput=1
\documentclass[10pt, logo, copyright]{nvidiatechreport}
\usepackage{microtype}
\usepackage{multirow}

\title{NV-Generate-Body: 3D Latent Diffusion Model for Controllable Anatomy, Tumor, and Demographic Body Mask Synthesis}

\author[1]{Can Zhao$^*$}
\author[1]{Pengfei Guo$^*$}
\author[1]{Dong Yang}
\author[1]{Yufan He}
\author[1]{Yucheng Tang}
\author[1]{Ahmed Harouni}
\author[2]{Benjamin Simon}
\author[3]{Mason Belue}
\author[2]{Stephanie Harmon}
\author[2]{Baris Turkbey}
\author[1]{Daguang Xu}

\correspondingauthor{canz@nvidia.com}

\affil[1]{NVIDIA}
\affil[2]{National Institutes of Health}
\affil[3]{University of Arkansas for Medical Sciences}

\begin{document}

\begin{abstract}
Curating large labeled CT datasets is expensive and constrained by patient privacy.
We present NV-Generate-BodyMask, a 3D latent diffusion model that synthesizes
whole-body CT segmentation masks with precise, joint control over organ anatomy,
tumor burden, and patient demographics.
The model operates in the compressed latent space of a frozen mask autoencoder,
generating $256^3$ volumetric label maps conditioned on a 19-dimensional vector
via Adaptive Group Normalization and cross-attention.
A size-adaptive ROI loss with spatial halo weighting and a paired tumor injection
pipeline enable accurate tumor size conditioning despite the scarcity of annotated
tumor cases.
Trained on 8,571 masks across 26 cohorts spanning 132 labels, the model achieves
Spearman $\rho = 0.70$--$0.92$ for organ size, 85--100\% tumor recall for lung,
pancreatic, colon, and bone lesions, and strong demographic alignment
(BMI $\rho = 0.78$, sex-conditioned perfect prostate control).\footnotemark
\end{abstract}

\maketitle
{%
  \renewcommand{\thefootnote}{\fnsymbol{footnote}}%
  \setcounter{footnote}{1}%
  \footnotetext{Equal contribution.}%
}%
\footnotetext{Code: \url{https://github.com/NVIDIA-Medtech/NV-Generate-CTMR}\\
\hspace*{1.8em}Model weights: \url{https://huggingface.co/nvidia/NV-Generate-CT/tree/main/models}}

\section{Introduction}

No existing generative system can synthesize whole-body CT segmentation masks with
joint, fine-grained control over organ sizes, tumor burden, and patient demographics.
Annotated 3D CT datasets are expensive to acquire, difficult to annotate at the
full-body scale, and restricted by patient privacy, making controllable synthetic
data a critical enabler for segmentation model training.
Yet current generative approaches remain too coarse: they operate on images rather
than structured labels, condition on coarse scene-level attributes, or rely on
fixed anatomical templates.

Existing generative approaches address pieces of this problem in isolation.
Models for individual organ or lesion synthesis~\cite{organ_synth_pinaya2022,organ_synth_chen2021,tumor_synth_jin2021,tumor_synth_hu2023}
lack whole-body scope.
Full-body CT synthesis models~\cite{ct_synth_maisi2024,ct_synth_maisi_v2,ct_synth_rombach2022}
operate in image space and do not expose explicit controls over anatomy or demographics.
Atlas-based mask generation~\cite{synthseg2023} provides diversity in appearance but
not in anatomy or pathology.
No existing system offers joint control over organ sizes, multiple tumor types,
and patient demographics within a unified 3D generative model.

We make the following contributions:
\begin{itemize}
  \item A 3D latent diffusion model operating in compressed mask latent space,
    conditioned via Adaptive Group Normalization (AdaGN) and cross-attention on a
    19-dimensional vector covering organ size quantiles, five tumor types, and
    four demographic variables (organ, tumor, and demographic slots described in
    Section~\ref{sec:problem}).
  \item A synthetic tumor injection pipeline with paired-patient crop selection and
    quantile-normalized conditioning, enabling the model to learn tumor size and
    location conditioning from a small annotated tumor library.
  \item A size-adaptive ROI loss with three-ring spatial halo weighting that prevents
    both under-generation of rare structures and boundary oversizing.
  \item A two-phase training scheme on 8,571 CT masks across 26 cohorts, delivering
    Spearman $\rho = 0.70$--$0.92$ for organ size, $85$--$100\%$ tumor recall, and
    precise demographic alignment at inference.
\end{itemize}

\section{Related Work}

\paragraph{Medical image synthesis with latent diffusion models.}
LDMs have enabled tractable 3D medical image synthesis but operate on intensity images and
lack explicit control over anatomy or pathology~\cite{ct_synth_rombach2022}.
MAISI~\cite{ct_synth_maisi2024} and MAISI-v2~\cite{ct_synth_maisi_v2} extend LDMs to full-body CT
synthesis at clinical resolution with coarse anatomy conditioning, and Pinaya et al.~\cite{organ_synth_pinaya2022} apply LDMs to brain MRI.
Earlier volumetric GAN approaches~\cite{hagan2022} demonstrated high-resolution 3D synthesis
but lack the fine-grained controllability of diffusion-based methods.
We instead generate structured label maps, which serve as a spatial layout that any paired
image synthesizer can condition on.

\paragraph{Segmentation mask synthesis and atlas-based methods.}
Prior work uses segmentation masks to condition image synthesis but does not generate
diverse masks from scratch with controllable anatomy.
SPADE~\cite{spade2019} synthesizes realistic images from fixed masks, demonstrating that
mask structure drives anatomical plausibility.
SynthSeg~\cite{synthseg2023} samples brain label maps from a generative model to diversify
MRI training data, but the underlying mask generator follows a fixed brain atlas and
cannot vary organ sizes or introduce pathology.
We close this gap by generating the mask itself, providing an upstream diversity signal
with precise, quantile-level control.

\paragraph{Tumor injection and copy-paste augmentation.}
Copy-paste augmentation can expand annotated tumor datasets but provides no mechanism for
conditioning on desired tumor size or quantile.
Hu et al.~\cite{tumor_synth_hu2023} and Chen et al.~\cite{chen2023towards} transplant real
tumor crops into healthy images and show that controlling size and host-organ context
improves generalization.
Jin et al.~\cite{tumor_synth_jin2021} address segmentation of gross tumor volumes.
Our approach extends copy-paste to the mask domain, using paired-patient crop selection
to preserve anatomical consistency and mapping the achieved tumor volume to a quantile
that the diffusion model learns to reproduce at inference.

\section{Method}

\subsection{Problem Formulation}
\label{sec:problem}

We formulate mask synthesis as learning the conditional distribution
$p(\mathbf{y} \mid \mathbf{c})$ over $256^3$ segmentation label maps, where
$\mathbf{c} \in \mathbb{R}^{19}$ jointly encodes organ sizes, tumor burden, and patient
demographics.
Specifically, $\mathbf{c}$ encodes:

\begin{itemize}
  \item \textbf{Slots [0--8]:} Per-organ size quantiles for 9 organs (liver, spleen, stomach,
    pancreas, right kidney, left kidney, lung, gallbladder, body), normalized to $[0,1]$ via
    per-organ empirical CDFs. A value of $-1$ indicates partial organ, organ absence,
    or unknown (e.g.\ organ out of field of view or size not annotated).
  \item \textbf{Slots [9--13]:} Per-tumor burden quantiles for 5 tumor types (lung tumor,
    pancreatic tumor, hepatic tumor, colon cancer primaries, bone lesion).  A value of $-1$
    indicates tumor absence.
  \item \textbf{Slots [14--18]:} Patient demographics (age, sex, weight, BMI, height).
    Continuous fields are min-max normalized to $[0,1]$ using fixed population ranges;
    sex is categorical (F\,=\,0, M\,=\,1); $-1$ indicates unknown.
    See Table~\ref{tab:demog_slots} for per-slot details.
\end{itemize}

\begin{table}[h]
\centering
\caption{Demographic conditioning slots [14--18]: field, encoding, and normalization range.}
\label{tab:demog_slots}
\begin{tabular}{clll}
\toprule
Slot & Field & Type & Range / Encoding \\
\midrule
14 & Age    & continuous & $[0,\,120]$\,yr, min-max $\to [0,1]$ \\
15 & Sex    & categorical & F\,=\,0,\ M\,=\,1,\ unknown\,=\,$-1$ \\
16 & Weight & continuous & $[0,\,200]$\,kg, min-max $\to [0,1]$ \\
17 & BMI    & continuous & $[0,\,75]$\,kg/m$^2$, min-max $\to [0,1]$ \\
18 & Height & continuous & $[0,\,200]$\,cm, min-max $\to [0,1]$ \\
\bottomrule
\end{tabular}
\end{table}

\subsection{Architecture}

\paragraph{Mask Autoencoder.}
A KL-regularized 3D convolutional autoencoder encodes
$256^3$ label maps in the 125-class VISTA label space~\cite{vista3d2024} into a
$64^3 \times 4$-channel latent representation, achieving $4\times$ spatial compression per
axis.
The decoder reconstructs per-class logits; the predicted label is the argmax.
The autoencoder is frozen throughout diffusion training.

\paragraph{Diffusion U-Net.}
The generative backbone is a 3D U-Net with Adaptive Group Normalization (AdaGN)~\cite{dhariwal2021diffusion}.
AdaGN modulates the scale and shift of each group normalization layer via a learned
projection of $\mathbf{c}$, enabling fine-grained per-layer conditioning on anatomy, tumor burden,
and demographics.
Cross-attention is applied at the two coarsest resolution levels.
A voxel-spacing embedding concatenated with the time embedding enables generation at arbitrary
isotropic spacings without retraining.
Table~\ref{tab:arch} summarises the key architectural differences from the baseline model.

\begin{table}[h]
\centering
\caption{Diffusion UNet architecture and conditioning: v1 vs.\ v2. \checkmark = present.}
\label{tab:arch}
\begin{tabular}{lcc}
\toprule
Component / Parameter & v1 & v2 \\
\midrule
Channel multipliers [64, 128, 256, 512]           & \checkmark & \checkmark \\
Residual blocks per level (2)                     & \checkmark & \checkmark \\
Self-attention (levels 3\,\&\,4, 32 heads)        & \checkmark & \checkmark \\
Cross-attention                                   & \checkmark & \checkmark \\
Flash attention                                   & \checkmark & \checkmark \\
Organ control: liver, pancreas, stomach,          &            &            \\
\quad colon, gallbladder                          & \checkmark & \checkmark \\
Tumor slots: lung, pancreatic, hepatic,           &            &            \\
\quad colon, bone                                 & \checkmark & \checkmark \\
\textbf{Voxel spacing}                            & $1.5^3$\,mm fixed & flexible \\
\textbf{Noise scheduler}                          & DDPM       & Rectified Flow \\
\textbf{AdaGN conditioning (dim 256)}             & ---        & \checkmark \\
\textbf{Organ control: spleen, left/right kidney, lung} & --- & \checkmark \\
\textbf{Sex, weight, BMI control}                  & ---        & \checkmark \\
\textbf{Classifier-free guidance (CFG)}           & ---        & \checkmark \\
\textbf{Training data augmentation}               & ---        & \checkmark \\
\textbf{Size-adaptive ROI loss}                   & ---        & \checkmark \\
\bottomrule
\end{tabular}
\end{table}

\paragraph{Noise Scheduler.}
We use Rectified Flow~\cite{liu2022rectified} with a linear noise schedule.
Inference runs 100 steps with classifier-free guidance (CFG)~\cite{ho2022cfg} at scale 2.0.
On a single NVIDIA H100 GPU, generating one $256^3$ mask takes approximately 4\,seconds.

\subsection{Datasets}

\paragraph{Segmentation corpus.}
We curate 8,571 CT cases from 26 public and institutional cohorts (Table~\ref{tab:cohorts}),
spanning thoracic, abdominal, and pelvic anatomy.
All cases are segmented to VISTA 132 labels.
Each case is pre-cropped to $256^3$ voxels at its native scanner spacing.

\begin{table}[h]
\centering
\caption{Training cohorts (top 10 by size).}
\label{tab:cohorts}
\begin{tabular}{lr}
\toprule
Cohort & Cases \\
\midrule
NLST & 2,447 \\
TCIA\_Colon & 1,141 \\
StonyBrook-CT & 1,103 \\
AbdomenCT-1K & 558 \\
NSCLC\_Radiogenomics & 531 \\
TotalSegmentatorV2 & 500 \\
LIDC & 338 \\
Lung-PET-CT-Dx & 330 \\
AMOS22 & 192 \\
Bone-NIH & 178 \\
16 additional cohorts & 553 \\
\midrule
\textbf{Total} & \textbf{8,571} \\
\bottomrule
\end{tabular}
\end{table}

\paragraph{Tumor library.}
A dedicated tumor crop library is built from cases with ground-truth tumor annotations.
Per-type volume statistics are summarized in Table~\ref{tab:tumorlibrary}.

\begin{table}[h]
\centering
\caption{Tumor library statistics.}
\label{tab:tumorlibrary}
\begin{tabular}{lrrrr}
\toprule
Tumor type & Crops & $p_{10}$ (mL) & $p_{50}$ (mL) & $p_{90}$ (mL) \\
\midrule
Lung tumor          & 63  & 1.3  & 5.2  & 55.6  \\
Pancreatic tumor    & 281 & 1.7  & 5.8  & 18.9  \\
Hepatic tumor       & 615 & 1.8  & 18.3 & 234.6 \\
Colon cancer        & 126 & 6.1  & 21.6 & 71.9  \\
Bone lesion         & 112 & 19.2 & 64.1 & 262.7 \\
\bottomrule
\end{tabular}
\end{table}

\paragraph{Demographics.}
Patient demographic metadata is available for a subset of cases and is summarized in
Table~\ref{tab:demog_stats}.
Demographic fields serve as conditioning signals during training and are not required at
inference; any unknown field is set to a null token.

\begin{table}[h]
\centering
\caption{Demographic coverage in the training set (total 8,571 cases).}
\label{tab:demog_stats}
\begin{tabular}{lrrrrl}
\toprule
Variable & $N$ & Coverage & Mean & Std & Range \\
\midrule
BMI (kg/m$^2$)  & 1,020 & 11.9\% & 29.3 & 6.3  & 18--63  \\
Weight (kg)     & 666   & 7.8\%  & 73.4 & 17.1 & 22--144 \\
Age (yr)        & 930   & 10.9\% & 64.1 & 12.0 & 19--87  \\
Sex             & 1,988 & 23.2\% & \multicolumn{3}{l}{1,197 M (60.2\%) / 791 F (39.8\%)} \\
\bottomrule
\end{tabular}
\end{table}

\subsection{Data Preparation}

\paragraph{Label space.}
We extend the 132-class VISTA label space~\cite{vista3d2024} with a \textbf{body envelope}
class that marks the entire foreground body silhouette and serves as the conditioning anchor
for the body-size quantile.
VISTA covers thorax, abdomen, pelvis, and skeleton; the body envelope is computed
automatically at preprocessing time (Section~\ref{sec:body_envelope}).
At inference, generated labels are remapped to the standard VISTA 132-class space via a
fixed look-up table, and the body envelope class is discarded.

\begin{figure}[t]
  \centering
  \includegraphics[width=\linewidth]{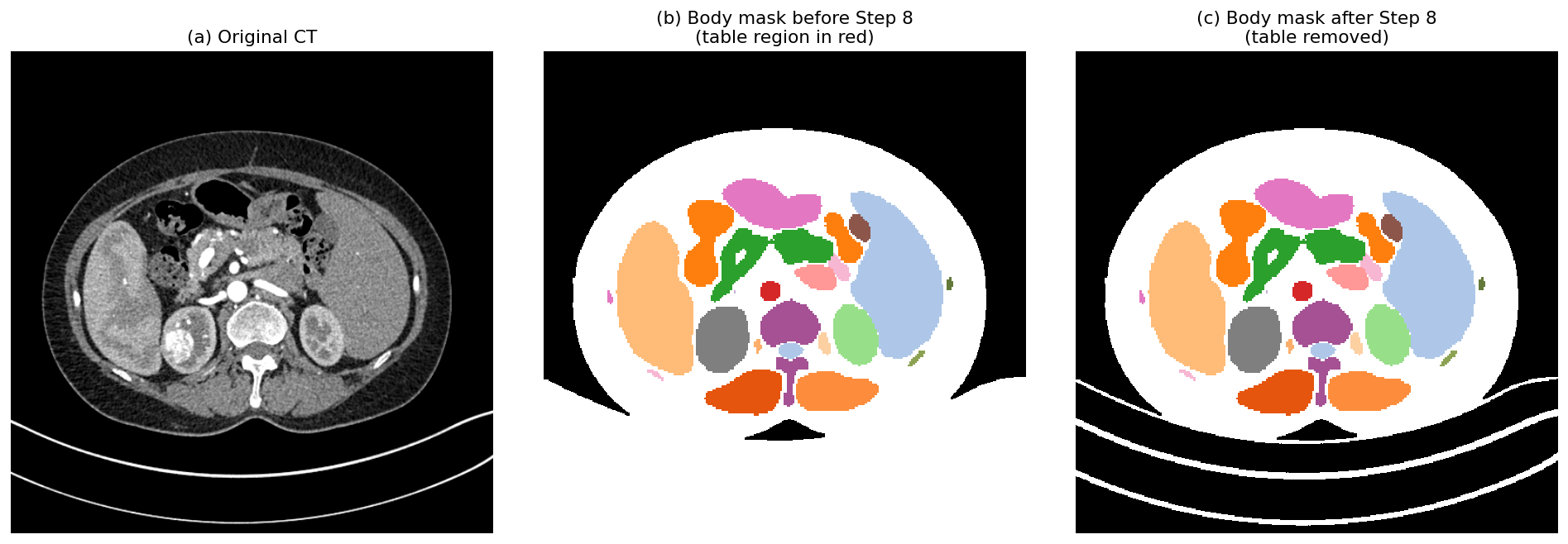}
  \caption{Body envelope construction on a case where three scanner table components
    account for 41\% of the initial body volume. \textbf{(a)}~Original CT axial slice.
    \textbf{(b)}~Body mask before table removal: all three components incorrectly included.
    \textbf{(c)}~Body mask after removal of all residual air-density regions inside the body.}
  \label{fig:body_envelope}
\end{figure}

\paragraph{Body envelope construction.}
\label{sec:body_envelope}
We construct the body envelope through a three-stage pipeline that robustly separates
patient tissue from the scanner bed under diverse table configurations.
Naive inversion of the air mask fails because the air gap between patient and table forms
a separate connected component from exterior air, causing the table to be included as body.
We construct the body envelope in three stages.
First, the exterior air region is identified as the largest connected component of
sub-air-threshold voxels ($\text{HU} < -800$), with organ-labeled voxels excluded to
prevent emphysematous lung from being absorbed into the exterior.
Second, the scanner bed is isolated by morphological erosion, largest-component selection,
and dilation, then removed.
Third, residual air-density regions remaining inside the body mask — arising from table
foam, side rails, or split table structures — are detected and removed regardless of
their number or spatial arrangement (Figure~\ref{fig:body_envelope}).
Internal cavities (bowel gas, lung airways) are sealed by a closing operation before the
remaining unlabeled voxels are assigned the body envelope class.

\paragraph{Anatomy size measurement.}
For each training case, organ volumes are measured at native spacing and converted to
per-organ quantiles via pre-computed empirical CDFs to form the size conditioning inputs.

\subsection{Training Data Augmentation Pipeline}

\paragraph{Geometric augmentation.}
At training time we apply three geometric augmentations in sequence.
Random in-plane translation shifts the mask by a small number of voxels in the axial plane
without $z$ displacement, preventing anatomy from sliding off the superior/inferior boundary.
Voxel-size jitter independently scales each axis by a random factor, with the spacing
conditioning rescaled accordingly so the model always trains on physically consistent pairs.
After geometric augmentation, organ size quantiles are recomputed from the deformed mask so
the model never trains on stale conditioning targets.

\paragraph{Demographic jitter.}
Continuous demographic fields receive small random perturbations before normalization,
reducing overfitting to the precise reported values in the training cohorts.

\paragraph{Synthetic tumor injection.}
Half of all training samples receive a synthetic tumor graft.
We sample a tumor type and select a crop from the library using \emph{paired boost}: crops
from the same source patient are assigned higher selection probability, preserving anatomical
consistency between the host mask and the injected tumor.
The crop is resampled to the host spacing and placed within the target organ mask by
maximizing organ-boundary overlap.
The achieved graft volume is mapped to a conditioning quantile via an empirical CDF of
real patient tumor burdens, which is written to the corresponding tumor slot.

\paragraph{Class-balancing and CFG dropout.}
Rare and anatomically small organs are up-sampled during training to compensate for
class imbalance.
With probability 0.1, the entire conditioning vector is set to the null token,
enabling classifier-free guidance at inference.

\subsection{Loss}
\label{sec:roi_loss}

We introduce a size-adaptive ROI loss that amplifies the training signal for small or
rare anatomical structures, which the vanilla diffusion MSE objective systematically
under-weights.
The loss augments the base diffusion objective with per-anatomy ROI terms:
\begin{equation}
\mathcal{L} = \mathcal{L}_{\text{diffusion}}
  + \sum_{r} w_r(q_r) \cdot \mathcal{L}_{\text{MSE}}(\text{ROI}_r),
\end{equation}
where the effective weight for ROI region $r$ at conditioning quantile $q$ is:
\begin{equation}
w_r(q) = 1 + \bigl(w_r^{\text{base}} - 1\bigr)|1 - q|.
\end{equation}
Small tumors (low $q$) receive the full base weight while large tumors
(high $q$) receive proportionally less amplification, preventing the loss from
over-focusing on extreme cases.

A spatial halo around each ROI mask is essential for accurate size control, as a high
per-voxel weight applied only to the tumor core creates no gradient penalty for predicting
spurious tumor voxels just outside the true boundary, causing systematic oversizing.
We use a three-ring halo with geometrically decaying weights, penalising boundary errors
and encouraging precise size and boundary placement.
ROI base weights $w_r^{\text{base}}$ are listed in Table~\ref{tab:roi_weights}.

\begin{table}[h]
\centering
\caption{ROI loss base weights $w_r^{\text{base}}$.}
\label{tab:roi_weights}
\begin{tabular}{lc}
\toprule
Region & $w_r^{\text{base}}$ \\
\midrule
Lung tumor          & 150 \\
Pancreatic tumor    & 30  \\
Colon cancer        & 30  \\
Bone lesion         & 1.8 \\
Hepatic tumor       & 2   \\
Pancreas (organ)    & 3   \\
Gallbladder (organ) & 3   \\
\bottomrule
\end{tabular}
\end{table}

\subsection{Training Protocol}
\paragraph{Mask Autoencoder.}
The autoencoder is trained on $256^3$ segmentation crops.
Integer label maps are encoded as binary bit planes and compressed to a four-channel latent
representation via an encoder-decoder with residual blocks at three spatial scales.
Training minimizes a combined Dice–cross-entropy reconstruction loss with KL regularization
using Adam and a cosine learning rate schedule with linear warm-up.

\paragraph{Diffusion U-Net.}
Training proceeds in two phases.
In Phase 1, the network is trained on the organ size quantile slots alone under a cosine
learning rate schedule, establishing anatomically coherent shape and size priors for all
nine organs.
In Phase 2, starting from the Phase 1 checkpoint, tumor injection augmentation, the full
19-dimensional conditioning vector, and the ROI loss are activated, with a constant
learning rate.
Both phases use mixed-precision training; Phase 2 trains on 16 NVIDIA H100 GPUs with an
effective batch size of 176.

\section{Experiments}

\subsection{Evaluation Protocol}

We evaluate condition-alignment via a sweep protocol: for each anatomy or tumor type, we
generate $N = 50$ cases per conditioning quantile $q \in \{0.1, 0.2, \ldots, 0.9\}$ at
CFG $\in \{0, 1, 2\}$ using 30 RF steps at 1.5\,mm isotropic.
Tumor recall is the fraction of generated cases where tumor volume exceeds 1\,mL.

\paragraph{Spearman $\rho$.}
Size calibration is quantified by Spearman's rank-order correlation coefficient $\rho \in [-1, +1]$
between the requested conditioning quantile and the generated organ or tumor volume
(restricted to cases where the tumor is present, for tumor types).
A value of $\rho = 1$ indicates that the generated volume perfectly tracks the requested
quantile in rank order; $\rho = 0$ means no monotonic relationship; $\rho < 0$ means
the model inverts the requested ordering.
Higher $\rho$ is better.

Demographic alignment is evaluated by sweeping BMI $\in \{20, 25, 30, 35, 40\}$ and
weight $\in \{50, 65, 80, 95, 110\}$\,kg, measuring Spearman $\rho$ against total foreground
body volume.
Sex control is evaluated via prostate presence rate for male vs.\ female conditioning.

\subsection{Baselines}

We compare against a v1 baseline---an earlier mask diffusion model trained without AdaGN
conditioning, tumor injection, or demographic inputs---to isolate the contribution of each
component introduced in this work.

\section{Results}

\subsection{Organ Size Calibration}

\begin{figure}[t]
\centering
\includegraphics[width=\linewidth]{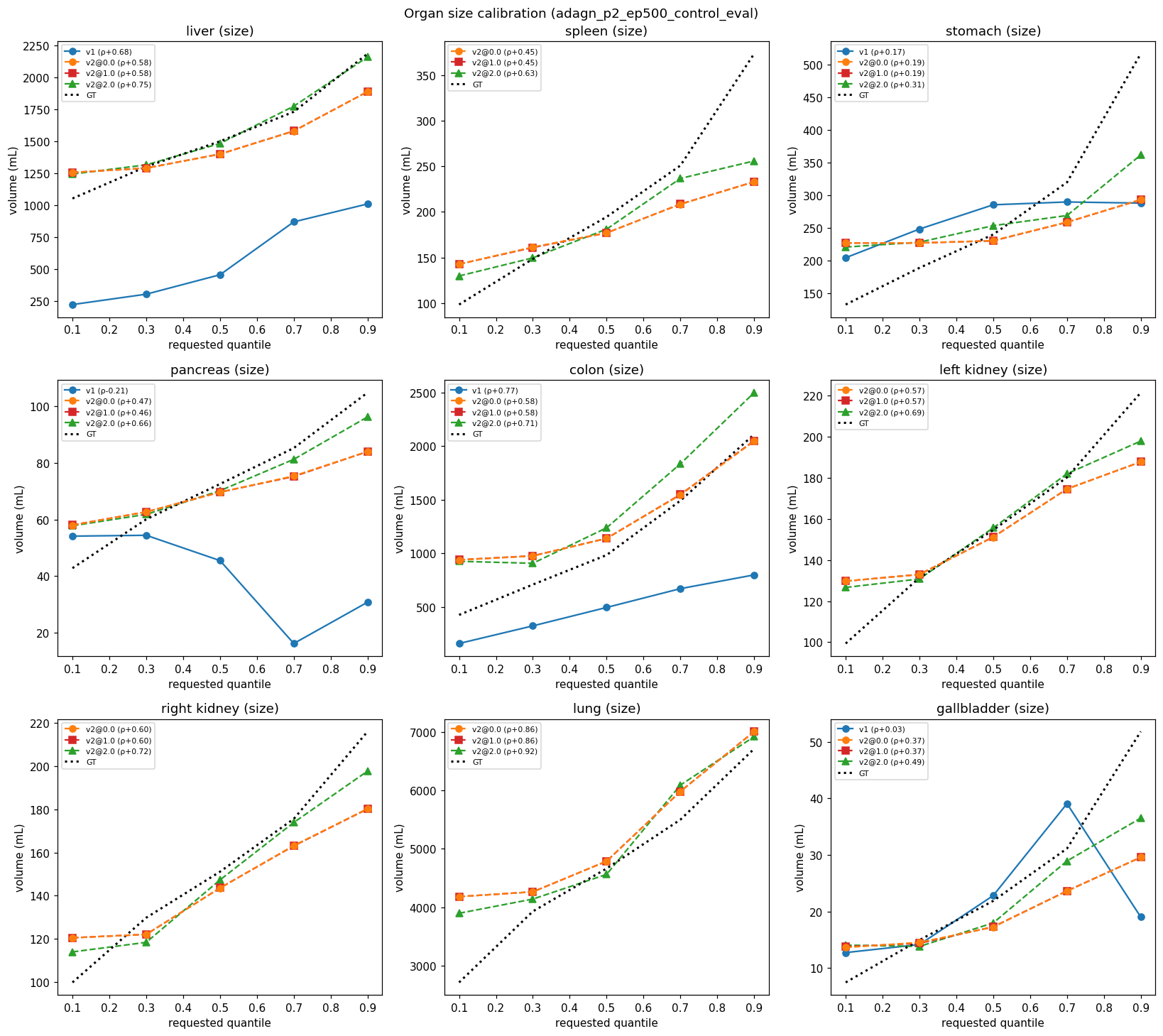}
\caption{Organ size calibration curves (CFG $\in \{0,1,2\}$).
  Each subplot shows generated volume (mL) vs.\ requested conditioning quantile,
  with the real-data reference overlaid. Spearman $\rho$ is reported per curve.}
\label{fig:organ_curves}
\end{figure}

AdaGN conditioning with the full augmentation pipeline delivers substantial improvements
in organ size calibration over the baseline across all nine organs
(Figure~\ref{fig:organ_curves}, Table~\ref{tab:organ}).
Lung ($\rho = 0.92$) and spleen ($\rho = 0.80$) benefit most, as these large and
variable organs produce clear size signals in the latent space.
Stomach ($\rho = 0.63$) and gallbladder ($\rho = 0.49$) remain more challenging: the stomach
deforms with filling state and the gallbladder is small relative to the $64^3$ latent grid.
Increasing CFG from 0 to 2 consistently improves calibration across all organs,
confirming that guidance amplifies the size conditioning signal at inference.

\begin{table}[h]
\centering
\caption{Organ size calibration (Spearman $\rho$, CFG\,=\,2). $\uparrow$ higher is better.}
\label{tab:organ}
\begin{tabular}{lcc}
\toprule
Organ & v1 & v2 \\
\midrule
Liver        & 0.68 & \textbf{0.75} \\
Spleen       & 0.49 & \textbf{0.80} \\
Stomach      & 0.17 & \textbf{0.63} \\
Pancreas     & 0.21 & \textbf{0.47} \\
Left kidney  & ---  & \textbf{0.69} \\
Right kidney & ---  & \textbf{0.72} \\
Lung         & ---  & \textbf{0.92} \\
Gallbladder  & 0.03 & \textbf{0.49} \\
\bottomrule
\end{tabular}
\end{table}

\subsection{Tumor Control}

\begin{figure}[t]
\centering
\includegraphics[width=\linewidth]{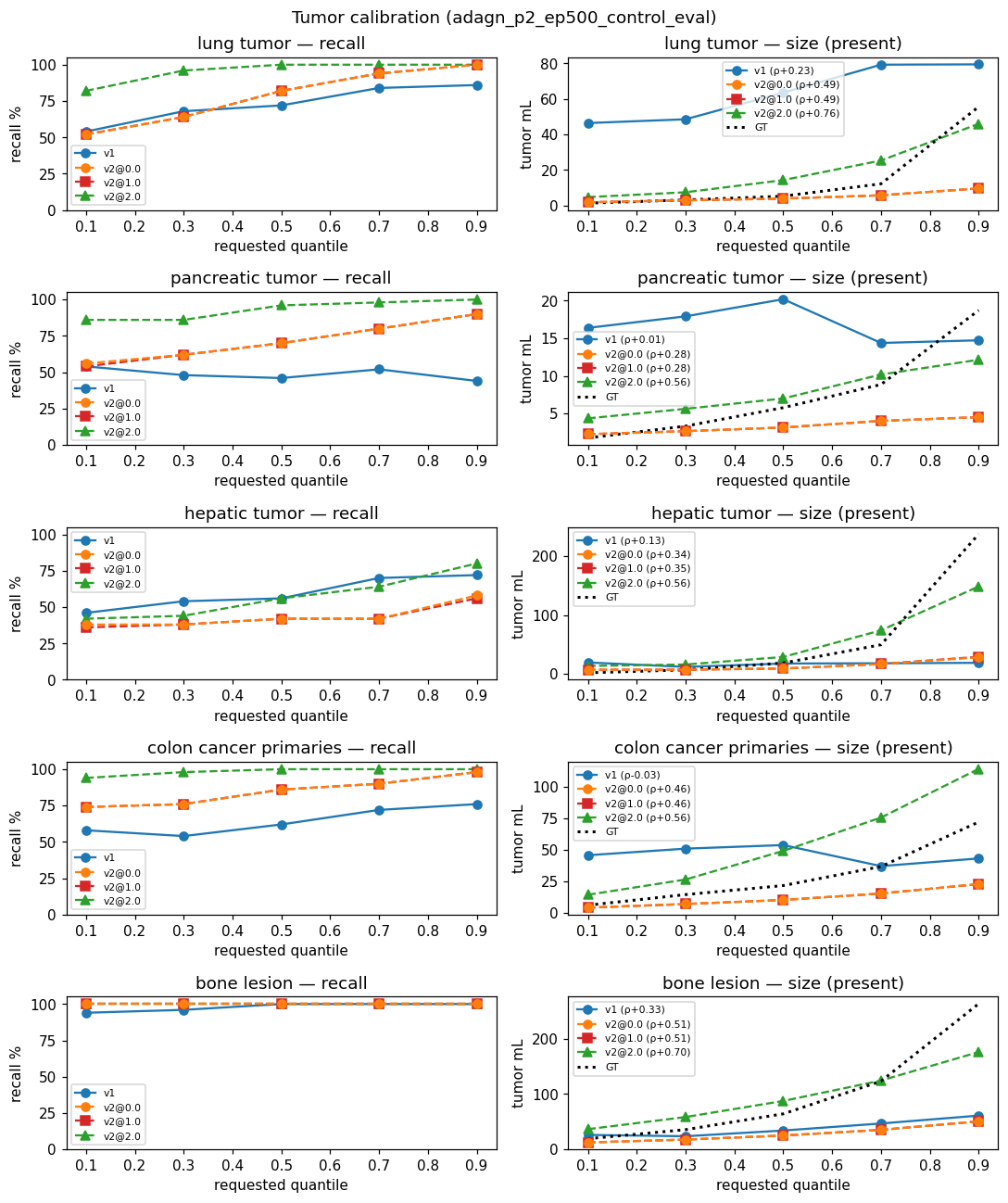}
\caption{Tumor recall (left) and size calibration (right) curves (CFG $\in \{0,1,2\}$).
  Left column: recall rate vs.\ requested conditioning quantile; GT is 100\% at all quantiles.
  Right column: mean generated tumor size (mL) vs.\ quantile among present cases,
  with the GT reference overlaid.
  Spearman $\rho$ is reported per curve.}
\label{fig:tumor_curves}
\end{figure}

Four of five tumor types reach $\geq 85\%$ recall across all conditioning quantiles at
CFG\,=\,2, validating the paired injection pipeline as a reliable source of tumor
conditioning signal (Figure~\ref{fig:tumor_curves}, Table~\ref{tab:tumor}).
The size-adaptive ROI loss is the decisive factor for lung tumor calibration ($\rho = 0.76$):
the diffuse, irregular morphology of lung tumors requires a base weight of 150 to produce a
measurable loss gradient relative to the surrounding lung parenchyma.
Without the three-ring halo, this high weight would produce boundary oversizing; the halo
redistributes the gradient to penalise out-of-boundary predictions (Section~\ref{sec:roi_loss}).

Hepatic tumor recall at CFG\,=\,0/1 is lower (35--45\%) than other tumor types, improving
to 60--80\% at CFG\,=\,2.
This gap reflects the structural challenge: the hepatic tumor is a small sub-label embedded
within a dominant liver region, and the model must simultaneously generate the liver mask and
place a distinct tumor sub-region inside it.
The low-quantile size signal is easily suppressed by the liver label's much larger gradient
contribution.
Colon cancer size is systematically overestimated at low quantiles ($\sim$14\,mL generated
vs.\ 6\,mL GT at $q{=}0.1$), suggesting that further tuning of the halo decay rate
and ROI base weight for this tumor type remains an avenue for improvement.

\begin{table}[h]
\centering
\caption{Tumor recall and size calibration. $\uparrow$ higher is better.
  v1 uses CFG\,=\,0 (no CFG); v2 uses CFG\,=\,2.}
\label{tab:tumor}
\begin{tabular}{lcccccc}
\toprule
 & \multicolumn{2}{c}{Recall ($q{=}0.1$)} & \multicolumn{2}{c}{Recall ($q{=}0.9$)} & \multicolumn{2}{c}{Size $\rho$} \\
\cmidrule(lr){2-3}\cmidrule(lr){4-5}\cmidrule(lr){6-7}
Tumor type & v1 & v2 & v1 & v2 & v1 & v2 \\
\midrule
Lung tumor       & 54\% & \textbf{82\%}  & 86\%  & \textbf{100\%} & 0.23 & \textbf{0.76} \\
Pancreatic tumor & 54\% & \textbf{86\%}  & 44\%  & \textbf{100\%} & 0.01 & \textbf{0.56} \\
Hepatic tumor    & \textbf{46\%} & 42\%  & 72\%  & \textbf{80\%}  & 0.13 & \textbf{0.56} \\
Colon cancer     & 58\% & \textbf{94\%}  & 76\%  & \textbf{100\%} & $-$0.03 & \textbf{0.56} \\
Bone lesion      & 94\% & \textbf{100\%} & \textbf{100\%} & \textbf{100\%} & 0.33 & \textbf{0.70} \\
\bottomrule
\end{tabular}
\end{table}

\subsection{Demographic Alignment}

Table~\ref{tab:demog} summarises alignment metrics.
BMI and weight conditioning produce strong monotonic alignment with generated body volume
($\rho = 0.78$ and $0.85$ respectively), confirming that the demographic slots propagate
through the latent space to shape body habitus.
Sex conditioning achieves perfect prostate control: prostate presence reaches 100\% for
male conditioning and 0\% for female conditioning across all CFG values, demonstrating
that the model captures sex-specific anatomical structures precisely rather than merely
correlating sex with body size.

\begin{table}[h]
\centering
\caption{Demographic condition alignment. $\uparrow$ higher is better.}
\label{tab:demog}
\begin{tabular}{lcc}
\toprule
Metric & CFG\,=\,0 & CFG\,=\,2 \\
\midrule
BMI $\to$ body volume ($\rho$)    & 0.73 & \textbf{0.78} \\
Weight $\to$ body volume ($\rho$) & \textbf{0.85} & 0.81 \\
Male prostate recall              & \textbf{100\%} & \textbf{100\%} \\
Female prostate recall            & \textbf{0\%}   & \textbf{0\%}   \\
\bottomrule
\end{tabular}
\end{table}

\subsection{Qualitative Results}

\begin{figure}[t]
\centering
\includegraphics[width=\linewidth]{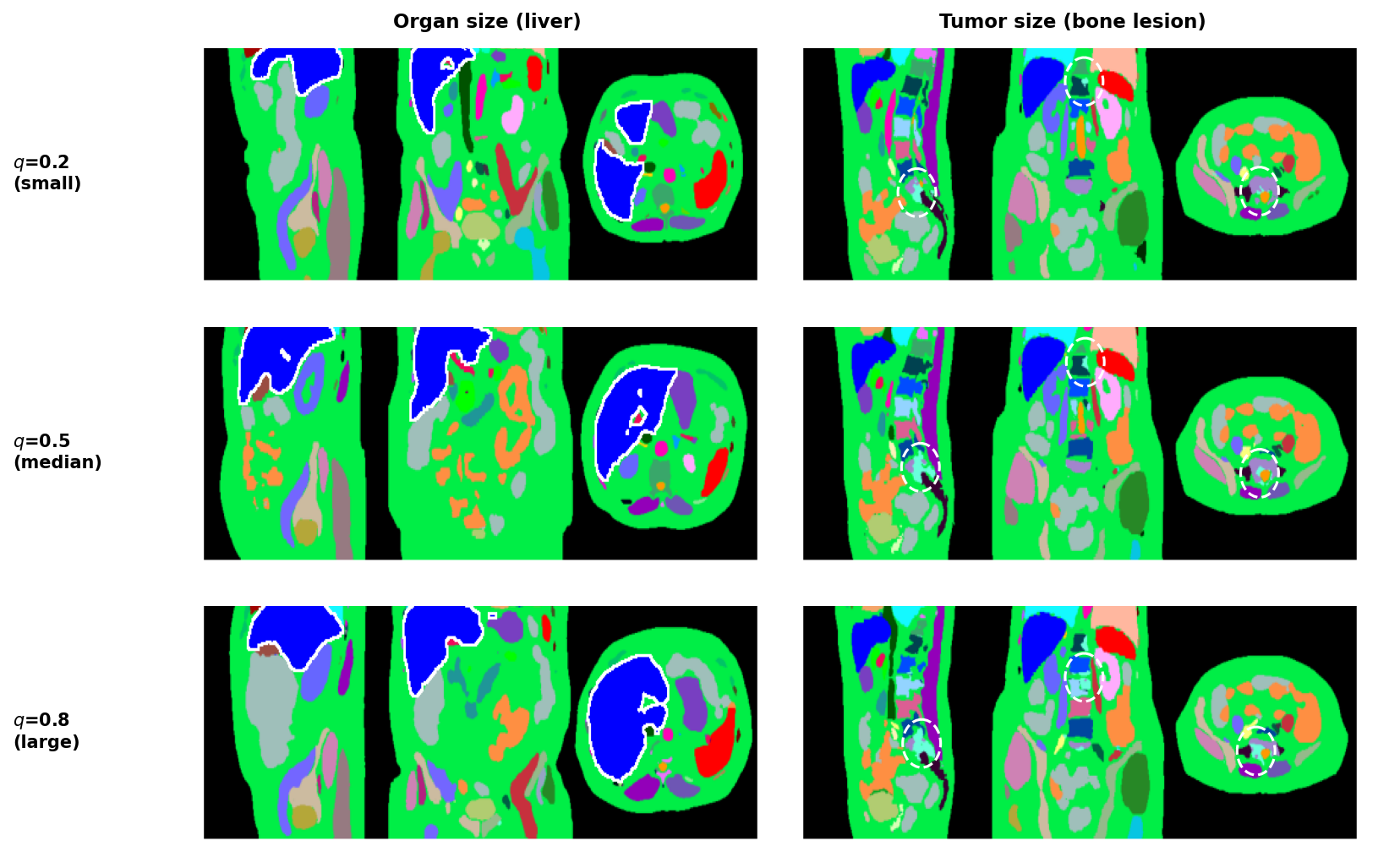}
\caption{Organ and tumor size conditioning.
  Each panel shows coronal, sagittal, and axial views resampled to isotropic spacing;
  organs are color-coded by label.
  Within each column, all three samples share the same random seed — only the
  requested size quantile differs.
  \textbf{Left column:} Liver at $q{=}0.2$ (small), $0.5$ (median), $0.8$ (large);
  white contour highlights the liver.
  \textbf{Right column:} Bone lesion at $q{=}0.2$, $0.5$, $0.8$;
  white dashed circle marks the lesion centroid.}
\label{fig:qualitative_size}
\end{figure}

\begin{figure}[t]
\centering
\includegraphics[width=\linewidth]{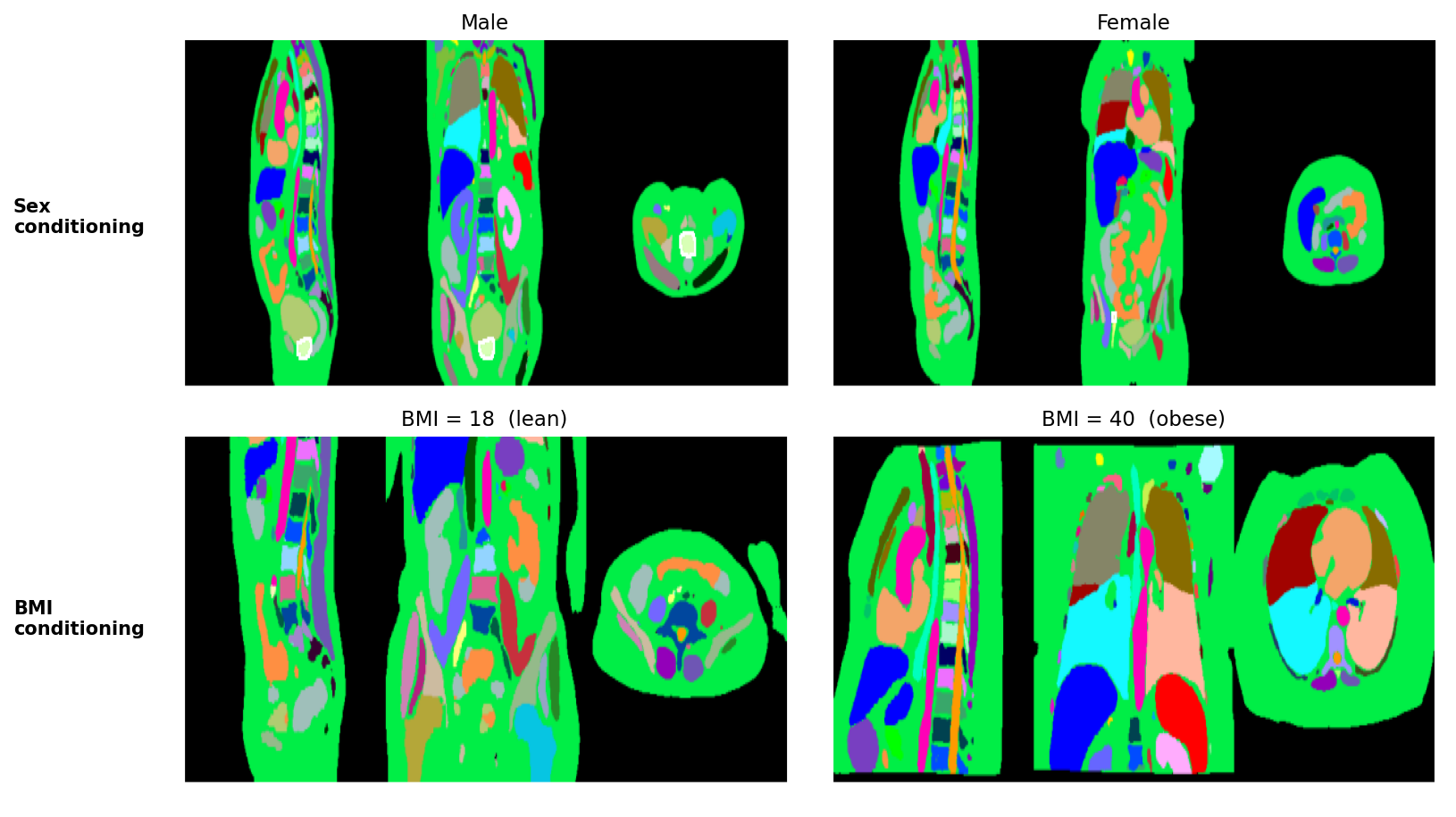}
\caption{Demographic conditioning. Each panel shows coronal, sagittal, and axial views.
  \textbf{Row 1:} Sex conditioning — Male (prostate highlighted with white contour) vs.\ Female
  (no prostate generated).
  \textbf{Row 2:} BMI conditioning — lean (BMI\,=\,18) vs.\ obese (BMI\,=\,40), showing
  visibly different body cross-sections.}
\label{fig:qualitative_demog}
\end{figure}

Figures~\ref{fig:qualitative_size} and~\ref{fig:qualitative_demog} show representative generated
masks for organ/tumor size control and demographic conditioning, respectively.
Liver size control produces visibly larger liver regions (outlined in white) as the quantile increases
from $q{=}0.2$ to $q{=}0.8$, while surrounding organ layouts remain anatomically plausible.
Bone lesion size control shows a clear progression from a small to a large lesion (white dashed circle marks the centroid).
Sex conditioning yields anatomically distinct configurations: a prostate (white outline) is present in
the male case and absent in the female case.
BMI conditioning produces a lean body habitus at BMI\,=\,18 and a larger cross-section with
increased adipose tissue at BMI\,=\,40.

\section{Conclusion}

NV-Generate-BodyMask is a 3D latent diffusion model that synthesizes whole-body CT
segmentation masks with joint control over organ anatomy, tumor burden, and patient demographics.
The 19-dimensional conditioning vector, AdaGN architecture, paired tumor injection pipeline,
and size-adaptive ROI loss combine to deliver Spearman $\rho = 0.70$--$0.92$ for organ size
calibration, 85--100\% tumor recall for lung, pancreatic, colon, and bone lesions, and
precise demographic control validated via BMI ($\rho = 0.78$), weight ($\rho = 0.85$), and
sex-conditioned perfect prostate generation.
The resulting masks can be paired with a downstream CT image synthesizer to generate
fully synthetic image--mask pairs, enabling on-demand production of demographically
stratified or pathology-enriched training data without real patient scans at inference time.

There are several limitations.
Hepatic tumor recall lags behind other tumor types (42\% at $q{=}0.1$) because the
hepatic tumor is a small sub-label embedded within a dominant liver region, making the
conditioning signal easy to suppress; dedicated presence losses or per-type ROI weight
tuning are natural remedies.
Colon cancer size is overestimated at low quantiles, suggesting the spatial halo decay
rate warrants further tuning for that tumor class.
Expanding the tumor library, extending demographic conditioning to underrepresented
subpopulations, and integrating mask generation directly with image synthesis are the
primary directions for future work.

\bibliographystyle{plain}
\bibliography{refs}

\end{document}